\pdfoutput=1

\documentclass[11pt]{article}

\usepackage[]{acl}

\usepackage{times}
\usepackage{latexsym}

\usepackage[T1]{fontenc}

\usepackage[utf8]{inputenc}

\usepackage{microtype}

\usepackage{graphicx}
\usepackage{xcolor}
\usepackage{adjustbox}

\usepackage[toc,page]{appendix}
\usepackage{multirow}
\usepackage{booktabs}
\usepackage{float}

\usepackage{tabularx}
\usepackage{soul}

\DeclareUnicodeCharacter{2212}{-}

\title{PreCogIIITH at HinglishEval : Leveraging Code-Mixing Metrics \& Language Model Embeddings To Estimate Code-Mix Quality}

\author{Prashant Kodali$^\dagger$ \qquad Tanmay Sachan$^\dagger$ \qquad Akshay Goindani$^\dagger$ \qquad Anmol Goel$^\dagger$ \\  {\bf Naman Ahuja}$^\dagger$ \qquad {\bf Manish Shrivastava}$^\dagger$ \qquad  {\bf Ponnurangam Kumaraguru}$^\dagger$\\
  $^\dagger$International Institute of Information Technology Hyderabad \\
  \texttt{\small \{prashant.kodali, tanmay.sachan, akshay.goindani, anmol.goel\}@research.iiit.ac.in}\\ \texttt{\small naman.ahuja@students.iiit.ac.in, \{m.shrivastava, pk.guru\}@iiit.ac.in}\\ 
  }

\begin{document}

\maketitle

\definecolor{codemix}{HTML}{FF9999} 
\definecolor{monolingual}{HTML}{E6FFDB}

\begin{abstract}
    Code-Mixing is a phenomenon of mixing two or more languages in a speech event and is prevalent in multilingual societies. Given the low-resource nature of Code-Mixing, machine generation of code-mixed text is a prevalent approach for data augmentation. However, evaluating the quality of such machine generated code-mixed text is an open problem. In our submission to HinglishEval, a shared-task collocated with INLG2022, we attempt to build models factors that impact the quality of synthetically generated code-mix text by predicting ratings for code-mix quality. HinglishEval Shared Task consists of two sub-tasks - a) Quality rating prediction); b) Disagreement prediction. We leverage popular code-mixed metrics and embeddings of multilingual large language models (MLLMs) as features, and train task specific MLP regression models. Our approach could not beat the baseline results. However, for Subtask-A our team ranked a close second on F-1 and Cohen's Kappa Score measures and first for Mean Squared Error measure.  For Subtask-B our approach ranked third for F1 score, and first for Mean Squared Error measure. Code of our submission can be accessed \href{https://github.com/prashantkodali/PreCogIIITH-HinglishEval-INLG-2022}{here}.
\end{abstract}

\section{Introduction}

Code-mixing\footnote{``Code-switching'' also refers to the phenomenon of mixing two or more languages and is often used interchangeably with code-mixing by the research community. Following the same convention, we use both terms interchangeably.} is a phenomenon where linguistic units from two or more languages are interspersed in a single utterance or a speech event and is common in multilingual communities. Due to increased penetration of the Internet and social media, code-mixing has become common and, at the same time, has posed challenges to automatic text processing pipelines~\citep{cetinoglu-etal-2016-challenges}. One such challenge is the dearth of naturally occurring code-mixed data. Data constraints have been the primary motive for researchers to leverage data augmentation and construct synthetically generated code-mixed corpora using monolingual parallel data as input. 

Synthetic code-mixed data generation, using monolingual parallel corpora,  is a non-trivial generation task. In the generated code-mixed sentence, one has to be careful about both the adequacy (preserving semantic content of monolingual sentence) and fluency (grammatical correctness). The task is further obscured by the fact that there is no single way of writing a code-mixed sentence.

\begin{figure*}[ht]
\includegraphics[width=\textwidth]{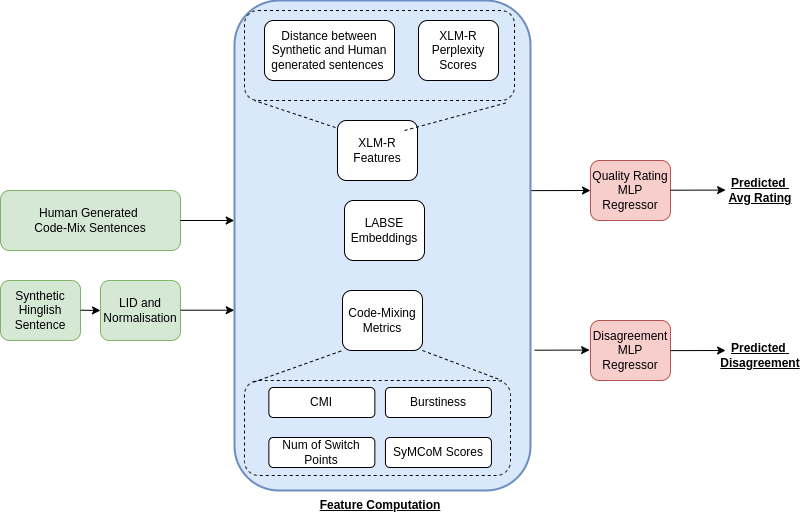}
\caption{System Architecture for predicting Average Quality Rating and Disagreement Rating of Synthetic Code-Mixed sentences. Human-generated code-mixed sentences are also used as input to our Regression model.}
\label{fig:blockdiag}
\vspace{-4mm}
\end{figure*}

In this work, we describe our approaches implemented in our submission to HinglishEval, a shared task co-located at INLG2022. HinglishEval is based on the HinGE dataset~\citep{srivastava-singh-2021-hinge}. HinGE is created in two phases: a) Human-generated Hinglish sentences:  at least two Hinglish sentences corresponding to the 1,976 English-Hindi sentence pairs; b) Synthetic Hinglish sentence generation and quality evaluation: generate Hinglish sentences using two rule-based algorithms, with human annotations for quality rating for each synthetically generated sentence. Two annotators rated each sentence on a scale of 1(low-quality) to 10 (high-quality). 

HinglishEval~\citep{srivastava-singh-2021-quality}, consists of two subtasks - a) Subtask-1 Quality rating prediction: For each synthetically generated sentence, predict the average (rounded-off) quality rating; b) Subtask-2 Disagreement prediction: predict the disagreement score (absolute difference between the two human ratings) for the synthetically generated sentence.

In our approach, we use a combination of code-mixed metrics and language model embeddings as features and train an MLP regressor for both the tasks. Rest of the paper is organised as follows: Section~\ref{system-overview} describes the features and models in detail as well as the experimental setup; Section ~\ref{results} covers the results of our experimentation; and we end with Section~\ref{discussion} discussing the implications, limitations, and future work.

\section{System Overview} \label{system-overview}
Figure~\ref{fig:blockdiag} shows the system architecture of our submission for both the Sub-Tasks. We describe the Pre-Processing involved, methodology for feature computation, and model architecture in subsequent sub-sections.

\subsection{Pre-Processing}
Before computing features on code-mixed sentences, we pre-process the sentences using CSNLI tool~\footnote{https://github.com/irshadbhat/csnli}. CSNLI computes the token-wise Language ID~(LID) and converts romanised Hindi tokens to native~(Devanagari) script, a step that is also known as Normalisation. LID tags are used to compute LID based code-mixing metrics. The normalised code-mixed sentences are useful in computing Multilingual Large Language Model (MLLM) features. MLLMs have been shown to perform better in downstream tasks when input code-mixed text is in normalised form~\citep{pires-etal-2019-multilingual}.

\begin{table*}[t]
\centering
\resizebox{\linewidth}{!}{%
\begin{tabular}{|c|c|c|c|} 
\hline
                                                                                                & \textbf{F1 Score}                                                        & \textbf{Cohen's Kappa}                                                    & \textbf{Mean Squared Error}                                        \\ 
\hline
\begin{tabular}[c]{@{}c@{}}\textbf{Sub-Task 1}\\\textbf{Quality rating prediction}\end{tabular} & \begin{tabular}[c]{@{}c@{}}0.25734 (2)\\$\Delta = 0.009$\end{tabular}  & \begin{tabular}[c]{@{}c@{}}0.09858 (2)\\$\Delta = 0.00064$\end{tabular} & \begin{tabular}[c]{@{}c@{}}2.00000 (1)\\$\Delta = 0$\end{tabular}  \\ 
\hline
\begin{tabular}[c]{@{}c@{}}\textbf{Sub-Task 2}\\\textbf{Disagreement prediction}\end{tabular}   & \begin{tabular}[c]{@{}c@{}}0.23523 (3)\\$\Delta = −0.02592$\end{tabular} & -                                                                         & \begin{tabular}[c]{@{}c@{}}3.00000 (1)\\$\Delta = 0$\end{tabular}  \\
\hline
\end{tabular}
}
\caption{Performance Measures of our system for individual Sub-Tasks. Values in the bracket show position of our system in the task leaderboard. $\Delta$ is indicating the difference between the top-performing system for the sub-task and our system.}
\label{tab:results}
\end{table*}

\subsection{Features}

Features used in our regression model can be broadly categorised as a) Code-Mixing Metrics as features, b) MLLM based Features.
    
    \begin{enumerate}
        \item Code-Mixing Metrics: \cite{Guzmn2017MetricsFM,gamback-das-2016-comparing} proposed multiple Language ID based metrics which are used to compare code-mix corpora. However, such measures fail to capture syntactic variety in code-mixing, and to overcome this limitation we utilise SyMCoM measures proposed by \citep{kodali-etal-2022-symcom}. We use the en-hi code-mix PoS tagger released by authors to compute PoS tags based on which SyMCoM scores are computed. For syntactic code-mix measures, we use SyMCoM scores for each PoS tag (Eq~\ref{SyMCoM}), and sentence level scores (Eq~\ref{SyMCoM-avg}). For Eq~\ref{SyMCoM} \& \ref{SyMCoM-avg}, $SU$ is a POS tag, and $L_1$ and $L_2$ are languages that are mixed. We use the following LID based code-mixing measures:
        \begin{itemize}
            \item Code-Mixing Index (CMI) as described in Eq.~\ref{cmi_eq}, where $N$ is the total number of languages mixed, $w_i$ is the number of words present from $i^{th}$ language, $n$ is the total number of tokens, and $u$ is the number of tokens given other tags.
            \item Number of Switch Points: number of times the language is switched within a sentence 
            \item Burstiness, as described in Eq.~\ref{burstiness_eq}, where $\sigma_{t}$ denotes the standard deviation of the language spans
and $m_{t}$ the mean of the language spans. Burstiness captures the periodicity in the switch patterns, with periodic dispersion of switch points taking on burstiness values closer to -1, and sentences with less predictable patterns of switching take on values closer to 1. 
        \end{itemize}
        
Code-mix metrics are only computed for the synthetic code-mix sentences. Further, we scale normalised code-mixing metric based features.

\begin{equation}\label{cmi_eq}
    CMI = \frac{\sum_{i=1}^{N}(w_i) - max(w_i)}{n-u}
\end{equation}

\begin{equation}\label{burstiness_eq}
    Burstiness = \frac{\sigma_{t} - m_{t}}{\sigma_{t} + m_{t}}
\end{equation}

        \item MLLM Features: In recent years, Multilingual Large Language Models (MLLMs), such as XLM-R~\citep{conneau-etal-2020-unsupervised}, have performed well across semantic tasks and cross-lingual transfer, and have been the go-to methods in code-mixed settings as well~\citep{khanuja-etal-2020-gluecos}. 
        We utilise embeddings from two models - XLM-R and LABSE~\citep{feng-etal-2022-language}. We compute the pseudo-log-likelihood(PPL) scores proposed by~\cite{salazar-etal-2020-masked}, which are akin to perplexity scores of conventional LMs. In our model, PPL scores are computed for both synthetic code-mixed sentences as well as human generated code-mixed sentences, and delta between the two PPL scores is considered as a feature. 
        
        We use LABSE model to compute sentence embeddings which are used as features. We compute LABSE embeddings for Hindi, English monolingual sentences, and synthetic code-mixed sentence. The intuition behind using features from two different LMs was to improve the discriminative power of the model.    
    \end{enumerate}
All the aforementioned features are concatenated resulting in a vector of dimension 2,385, and these features are used to train task-specific models.      

\begin{equation} \label{SyMCoM}
SyMCoM_{SU} = \frac{(Count_{SU_{L1}}) - (Count_{SU_{L2}})}{\sum_{i = 1}^{2} {Count_{SU_{Li}}}}
\end{equation}

\begin{equation} \label{SyMCoM-avg}
SyMCoM_{sent} = \sum_{SU} {\frac{Count_{SU}}{len}} \times |SyMCoM_{SU}|
\end{equation}

\subsection{Models}
We experimented with various models such as - Linear Regression, MLP Regressor, and XGBoost with the combination of the above features. In the Validation phase, we noticed that the MLP outperformed all the other models. For the test phase we used only the MLP Regressor models. We train task-specific MLP Regression models using the same features, and rely on them to learn the complex function to predict the task-specific values in the same feature space.

We use the Sklearn library~\citep{scikit-learn} to implement the MLP Regressor models. We implement the MLP with three hidden layers - consisting of 1000, 100, and 10 neurons, respectively, paired with ReLU activation functions, an Adam optimizer, adaptive learning rate, and a default batch size and number of epochs. We could do only a limited hyper-parameter search, and a more structured and comprehensive hyper-parameter search could lead to further improvement in the model's performance. Our hyperparameter search space for learning rate was {0.01, 0.001, 0.0001}, and for hidden layer dimensions search space was \{10,100,1000\}.

\section{Results} \label{results}

The scores for the Sub-Tasks achieved by our model are given in Table \ref{tab:results}. For Sub-Task 1, we achieved rank 2 on the leaderboard for F1-Score and Cohen's Kappa, while a rank of 1 for Mean Squared Error (tied with the baseline model).

For Sub-Task 2, we beat the baseline model and achieved rank three on the leaderboard for F1-Score and one for Mean Squared Error (tied with the models achieving ranks 1 and 2).

For Sub-Task 1, our system is closest to the baseline model, as none of the competing models would beat the baseline model's performance. We hypothesize that the low-performance scores can be attributed to the task's hardness and the data's size. 

As noted earlier, a comprehensive and structured hyper-parameter search will likely improve the results. Because of the very low delta between our system and the best performing system, hyper-parameter tuning could be crucial to surpassing the baseline models.

\section{Discussion} \label{discussion}
In this work, we propose a system to predict the Quality and Disagreement scores given code-mixed sentences and their monolingual counterparts. We leverage the combination of code-mixing metrics and MLLMs embeddings as features and train MLP regressor models. While our approach fails to beat the baseline/best performing system, the performance of our system is a close second or third and ranks first on MSE for both Sub-Tasks. Further hyper-parameter tuning can further improve the results. 

Even the best-performing systems/baselines have very low scores across performance measures, which can be attributed to the difficulty of the task at hand, and the subjectivity of annotators while rating a sentence on a scale of 1-10. The size of the dataset could also be a limitation for solving the task at hand. 

While MLP-based regressors are black boxes, having an explainable/interpretable model could help rank the features that impact the scores. In our system, an ablation study could help prune the feature space and identify the kind of features that are useful in rating prediction, and such features could be augmented. We leave these pursuits as part of our future work. 



\bibliography{anthology,custom}
\bibliographystyle{acl_natbib}




\end{document}